\documentclass{article}

\usepackage[preprint]{neurips_2026}

\usepackage[utf8]{inputenc} 
\usepackage[T1]{fontenc}    
\usepackage{hyperref}       
\usepackage{url}            
\usepackage{booktabs}       
\usepackage{amsfonts}       
\usepackage{nicefrac}       
\usepackage{microtype}      
\usepackage{xcolor}         
\usepackage{graphicx}
\usepackage{booktabs}
\usepackage{multirow}
\usepackage{subcaption}
\usepackage[none]{hyphenat}

\title{PRISM: Progressive Reasoning through Iterative Slot Memory for Vision}

\author{Ziyu Wang \quad 
Shuangpeng Han \quad 
Mengmi Zhang\\
Deep NeuroCognition Lab, Nanyang Technological University, Singapore \\
Address correspondence to \texttt{mengmi.zhang@ntu.edu.sg}
}

\begin{document}

\maketitle

\begin{abstract}
Modern vision models process images in a single feed-forward pass, which limits their ability to recover missing evidence or refine uncertain representations under incomplete observations.
Inspired by the iterative nature of human perception, we introduce PRISM (\underline{P}rogressive \underline{R}easoning through \underline{I}terative \underline{S}lot \underline{M}emory), a pyramid vision architecture that reasons over images through iterative refinement. At a high level, PRISM groups visual features into object-centric representations, retrieves relevant patterns from a learned memory, and iteratively refines the representation to resolve ambiguity and recover missing information. This organize–recall–refine process operates recurrently across multiple scales, enabling progressive improvement of visual representations. Across standard vision tasks, including image classification, object detection, and semantic segmentation, PRISM achieves competitive performance while demonstrating improved robustness under incomplete observations such as occlusion. These results suggest that iterative reasoning with structured representations and memory is a promising direction for building more resilient and adaptive vision models. Source code and models will be released.

\end{abstract}

\section{Introduction}
Visual recognition systems \cite{vaswani2017attention, liu2021swin, wang2021pvt, wang2021pvtv2,shi2211unveiling,jia2025seeing,talbot2023tuned,cai2025learning} are expected to operate reliably in real-world environments, where observations are often incomplete due to occlusion or missing visual evidence. Although modern deep networks achieve impressive performance on clean benchmarks, their representations can become fragile when important visual cues are unavailable \cite{lin2025make,wang2024pose,zhang2020putting,bomatter2021pigs,liu2022reason}. A key limitation is their predominantly feed-forward nature: visual features are computed in a single pass and propagated through layers without revisiting or refining intermediate representations. As a result, incomplete information is compounded in deeper layers, leading to degraded recognition rather than progressive inference.

A long-standing goal in representation learning is to model scenes in terms of object-centric entities \cite{locatello2020slotattention, wu2023slotformer, wang2023object, han2024flow}, such as objects and parts, rather than undifferentiated pixel collections or dense feature tokens arranged on rigid grids in transformer-based feature maps. Object-centric representations improve robustness by aggregating information across regions belonging to the same entity, allowing partial observations to remain interpretable through structured object-level evidence and enabling inference of occluded regions from visible parts.

However, object-centric grouping alone is not sufficient. Methods such as slot attention \cite{wang2024pose,wang2023object,han2024flow} organize features into coherent entities, but remain fundamentally bottom-up: the resulting slots reflect only the available evidence and become incomplete or ambiguous under occlusion. Without access to prior knowledge, a partial slot with degraded visual information due to occlusion cannot reliably infer the full structure of an object or disambiguate between competing interpretations.

Conversely, relying on a prototype memory alone (e.g., VQ-VAE\cite{oord2017vqvae}) is also insufficient. When applied directly to dense feature tokens, the inputs are entangled mixtures of multiple objects and background, leading to ambiguous or noisy prototype matching. As a result, the learned codebook tends to capture blurred or mixed patterns rather than clean object-level structures, making retrieval unreliable for completion.

This motivates combining both components. A vector-quantized memory provides global priors in the form of prototypical patterns, enabling restoration of missing features via top-down modulation, while slot attention supplies the necessary structure by grouping tokens into object-centric representations that are suitable for prototype matching. Together, they play complementary roles: slot attention organizes visual evidence, and the memory module enables completion by grounding these representations in learned prototypes. This synergy allows the model to both organize and recover visual information, overcoming the limitations of feed-forward architectures that lack mechanisms to organize, recall, and refine representations under incomplete observations.


In parallel, biological vision exhibits recurrent dynamics across the cortical hierarchy (e.g., V1–V2–V4), with extensive behavioral and neurophysiological evidence showing that perception is iteratively refined through feedback and lateral interactions \cite{zhang2020putting,tang2018recurrent,kietzmann2019recurrence,kubilius2019brain,kar2021fast} Loosely inspired by neuroscientific and behavioral evidence from biological vision systems, we propose PRISM, a pyramid vision architecture that introduces progressive reasoning through iterative slot memory. 

\begin{figure}[t]
    \centering
    \includegraphics[width=\linewidth]{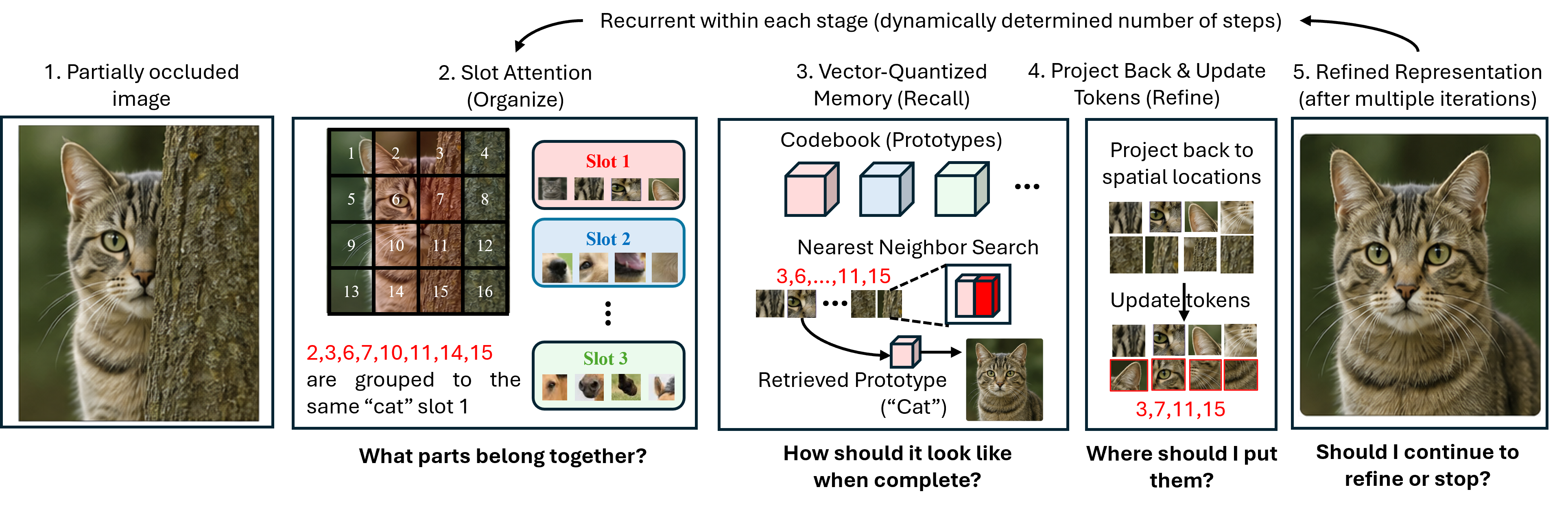}\vspace{-2mm}
    \caption{\textbf{Schematic of Progressive Reasoning through Iterative Slot Memory (PRISM).} During inference, the right half of the cat is occluded by a tree. PRISM performs progressive reasoning as follows: (1) visual tokens with similar semantics and structure are grouped into an object-centric “cat” slot; (2) the partial slot, degraded by occlusion, is matched to the nearest prototype in a learned memory during training, retrieving a complete cat pattern; (3) the retrieved prototype is projected back to the feature space to update tokens associated with the cat; and (4) this process is iterated for a dynamically determined number of steps until the representation becomes complete and coherent. This organize–recall–refine cycle is applied recurrently to progressively improve the representation.  
    }\vspace{-4mm}
    \label{fig:schematic}
\end{figure}

As illustrated in \textbf{Figure.~\ref{fig:schematic}}, PRISM first groups visual tokens into object-centric representations that capture the underlying visual structure. These slots are then mapped into a vector-quantized memory space, where a learned codebook stores prototypical slot patterns acquired during training.
When observations are incomplete, the corresponding slot representations may also be degraded. PRISM retrieves nearby prototypes from memory to restore these slots and projects the restored information back to their spatial locations to update token features.

This organize–recall–refine process is applied recurrently within each stage, enabling progressive refinement of representations over multiple iterations.
Moreover, these recurrent updates are adaptive within each stage: PRISM dynamically determines the number of refinement iterations, allowing more challenging inputs to receive additional reasoning steps while keeping simpler cases efficient. Together, these properties enable PRISM to treat visual inference not as a single feed-forward transformation, but as a recurrent process of progressively organizing visual evidence, recalling learned prototypes, and refining internal representations.


We evaluate PRISM on standard computer vision tasks, including image classification, object detection, and semantic segmentation. Across these settings, PRISM achieves competitive performance while demonstrating improved robustness under partial observations and missing visual evidence. These results highlight iterative slot memory as a general architectural principle for strengthening visual recognition beyond task-specific designs. The contributions of this paper are as follows:

\noindent \textbf{1.} We introduce a recoverable representation paradigm, where visual features are organized into object-centric abstractions and grounded in a learned prototype space, enabling reconstruction of coherent representations from incomplete observations.

\noindent \textbf{2.} We propose a recurrent, stage-wise adaptive refinement mechanism, where visual representations are iteratively updated through organize–recall–refine cycles, with the number of iterations dynamically determined and all tokens jointly updated at each step.

\noindent \textbf{3.} We show that prototype-based restoration combined with dynamic recurrent refinement provides a strong inductive bias for artificial vision systems, where representations are progressively refined rather than produced in a single feed-forward pass, yielding high efficiency and improved robustness under occlusion.

\section{Related Works} 

\textbf{Slot-Based and Object-Centric Representations.}
Prior work models scenes via object-centric representations by decomposing inputs into latent entities. Slot Attention and its variants group features into slots through iterative competition~\cite{locatello2020slotattention, wang2023object, wang2024pose, han2024flow}, and have been widely used in unsupervised object discovery and scene decomposition (e.g., MONet~\cite{burgess2019monet}, IODINE~\cite{greff2019iodine}). Extensions to video further introduce temporal consistency in object-centric representations (e.g., SAVi~\cite{kipf2022savi}, SlotFormer~\cite{wu2023slotformer}, VideoSAUR~\cite{zadaianchuk2023videosaur}).
However, these methods are typically standalone or auxiliary modules and are rarely integrated into high-performance multi-stage backbones for visual recognition trained on large-scale datasets, such as ImageNet \cite{deng2009imagenet}. Moreover, continuous slots remain sensitive to incomplete observations, leading to inconsistent representations. In contrast, PRISM embeds slot-based grouping within a pyramidal architecture and augments it with a discrete prototype memory, enabling stable and recoverable representations. It performs on par with competitive baselines on large-scale ImageNet recognition benchmarks and achieves superior performance on occluded ImageNet images without any prior exposure to occluded samples during training. 

\textbf{Discrete Representation Learning.}
Vector quantization maps continuous features to a finite set of learned prototypes. VQ-VAE~\cite{oord2017vqvae} and VQGAN~\cite{esser2021taming} demonstrate their effectiveness for image generative modeling, with subsequent works introducing improvements such as residual quantization~\cite{lee2022autoregressive}, hierarchical quantization~\cite{takida2024hqvae}, and lookup-free quantization~\cite{yu2024language}. These methods show that discrete codebooks provide compact and reusable visual representations. However, prior work has primarily used vector quantization for generative modeling or tokenization, rather than for restoration within discriminative visual recognition models. Moreover, applying VQ to dense token grids can entangle multiple objects, resulting in ambiguous prototype assignments. In contrast, PRISM performs quantization at the slot level, enabling prototype-based restoration of object-centric representations.

\textbf{Slot-Based Vision Models with Vector Quantization.}
Recent works integrate slot-based representations with discrete learning, including VQ-VFM-OCL~\cite{zhao2025vvo}, Semantic VQ~\cite{wu2024semanticvq}, and Slot-MLLM~\cite{chi2025slotmllm}. While effective for object-centric learning, world modeling, and multimodal tokenization, these approaches are not designed for general visual recognition tasks.
In contrast, PRISM leverages vector-quantized slots to recover degraded intermediate features within a general-purpose vision backbone, and applies this mechanism across multiple stages to enable progressive refinement.

\textbf{Recurrent and Adaptive Computation.}
Recurrent and adaptive computation methods enable models to dynamically adjust their computational depth, including ACT~\cite{graves2016adaptive}, Universal Transformers~\cite{dehghani2019universal}, and PonderNet~\cite{banino2021pondernet}. In vision, approaches such as A-ViT~\cite{yin2022avit} and depth-adaptive transformers~\cite{elbayad2020depthadaptive, liu2021fasterdepthadaptive} adjust inference depth based on input complexity. These methods primarily treat recurrence as variable-depth computation over dense feature representations. In contrast, PRISM grounds recurrence in prototype-based representation restoration, where each iteration explicitly refines features via vector-quantized memory, enabling structured and progressive recovery under incomplete observations. The superior performance of PRISM empirically supports this design.


\section{PRISM}

\begin{figure}[t]
    \centering
    \includegraphics[width=\linewidth]{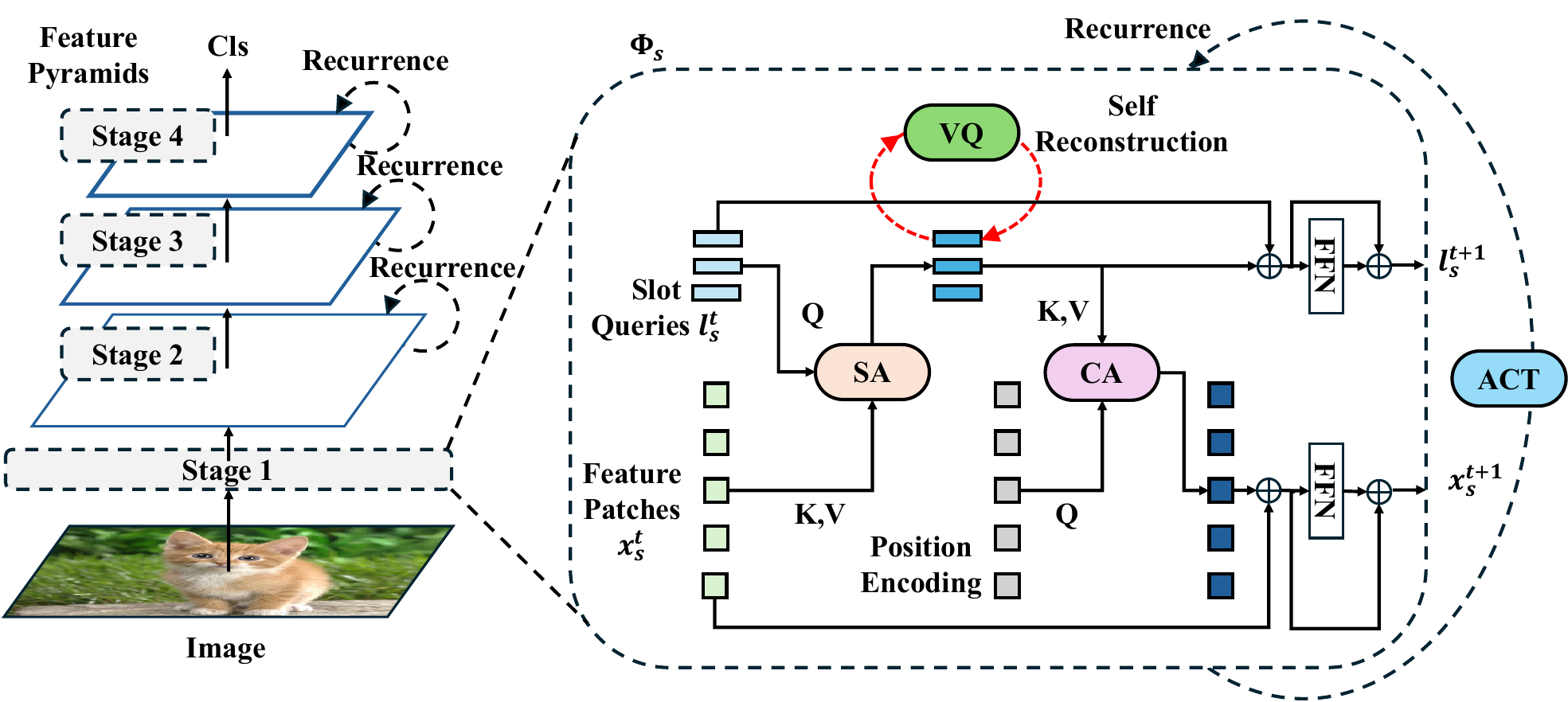}\vspace{-3mm}
    \caption{
    \textbf{PRISM Architecture.} PRISM consists of four stages in a feature pyramid, where each stage recurrently refines token features at different resolutions before passing them to the next stage, and the final representation is used for classification (Cls). Within each stage, feature tokens are first grouped into object-centric slots via Slot Attention (SA), which performs competitive grouping into a fixed number of slots to produce structured representations capturing distinct entities. These slot representations are grounded in a vector-quantized (VQ) memory trained with a self-reconstruction objective, where continuous features are mapped to a finite set of learned prototypes, yielding compact and reusable representations. The resulting prototypical slot features are then redistributed to their corresponding spatial locations through cross-attention (CA), where positional embeddings serve as queries (Q), enabling spatial information flow back to the feature map. The resulting feature maps are further processed by feed-forward networks (FFNs) with residual connections. We encapsulate these core operations within the $s$-th stage as a recurrent function denoted by $\Phi_s$ (the big dashed box). Each stage is executed recurrently via $\Phi_s$ to enable iterative refinement, with adaptive halting determining the number of recurrent steps through an Adaptive Computation Time (ACT) module based on input complexity.
    }\vspace{-4mm}
    \label{fig:prism_structure}
\end{figure}

PRISM is a four-stage vision pyramid architecture (\textbf{Figure.~\ref{fig:prism_structure}}), where each stage begins with a 2D convolutional layer that converts the input image (at the first stage) or the incoming feature map (at subsequent stages) into a sequence of visual tokens. Across stages, the spatial resolution is progressively reduced by a factor of two, forming a standard feature pyramid with resolutions of $\frac{1}{4}$, $\frac{1}{8}$, $\frac{1}{16}$, and $\frac{1}{32}$ of the input image, while the channel dimension increases accordingly, enabling PRISM to learn visual features at different levels of granularity.
Unlike conventional pyramid backbones such as PVT~\citep{wang2021pvt}, PVTv2~\citep{wang2021pvtv2}, and Swin Transformer~\citep{liu2021swin}, which stack multiple feed-forward blocks within each stage, PRISM employs a single block per stage and applies it recurrently.

As shown in \textbf{Figure.~\ref{fig:prism_structure}}, at the recurrent step $t$ of each stage $s$, PRISM takes feature tokens $\mathbf{x}_s^{t} \in \mathbb{R}^{N_s \times D_s}$, slot queries $\mathbf{l}_s^{t} \in \mathbb{R}^{K_s \times D_s}$, and positional embeddings $\mathbf{p}_s \in \mathbb{R}^{N_s \times D_s}$ as input, and builds on three core operations: slot attention (SA), vector quantization (VQ), and cross-attention (CA) to refine token representations. 
We denote these operations as $\mathrm{SA}(\cdot;\cdot;\cdot)$, $\mathrm{VQ}(\cdot;\cdot;\cdot)$, and $\mathrm{CA}(\cdot;\cdot;\cdot)$, respectively, and use them as fundamental building blocks throughout each stage of PRISM. Detailed mathematical formulations are provided in \textbf{Appendix~Section.~\ref{supp:preliminaries}}. 
We use $\Phi_s$ to denote the recurrent function that encapsulates all these core operations within the $s$-th stage. Each stage is executed recurrently via $\Phi_s$ to enable iterative token refinement, with adaptive halting determining the number of recurrent steps through an Adaptive Computation Time (ACT) module based on input complexity. See the mathematical formulation of $\mathrm{ACT}(\cdot;\cdot)$ in \textbf{Appendix~Section.~\ref{supp:preliminaries}}. 
Next, we introduce the three core modules, SA, VQ, and CA, within $\Phi_s$, as well as the ACT mechanism in detail.





\subsection{Grouping Semantically and Structurally Similar Tokens into Slot Entities}


PRISM performs competitive grouping over patch features into a fixed number of slots to obtain structured representations capturing distinct entities. Slot features $\mathbf{l}_s^{t}$ serve as queries, while position-aware patch features ($\mathbf{x}_s^{t} + \mathbf{p}_s$) and the original patch features $\mathbf{x}_s^{t}$ serve as keys and values:
\begin{equation}
\mathbf{v}_{s,\mathrm{grp}}^{t}
=
\mathrm{SA}\!\left(
W_{q,s}\mathbf{l}_s^{t},
W_{k,s}(\mathbf{x}_s^{t}+\mathbf{p}_s),
W_{v,s}\mathbf{x}_s^{t}
\right).
\end{equation}
$W_{q,s}, W_{k,s}$ and $W_{v,s}$ are learnable linear projection matrices.
Unlike classical Slot Attention \cite{locatello2020slotattention}, we incorporate positional modulation in the keys, encouraging slot features to capture not only semantic content but also spatial structure of object entities. Therefore, in addition to the grouped values $\mathbf{v}_{s,\mathrm{grp}}^{t}$, we also compute grouped keys $\mathbf{k}_{s,\mathrm{grp}}^{t}$, which are used in the subsequent redistribution step to place object-centric representations back at correct spatial locations.

Let $\mathbf{W}_{s,\mathrm{grp}}^{t}$ denote the attention weights measuring similarity between slot queries and position-aware keys produced by $\mathrm{SA}$ (see also the formulation of $W_{ij}$ in \textbf{Appendix~Equation.~\ref{equ:equAttSA}}). The grouped keys are obtained by weighting the position-aware keys with these attention weights:
\begin{equation}
\mathbf{k}_{s,\mathrm{grp}}^{t}
=
\mathbf{W}_{s,\mathrm{grp}}^{t} W_{k,s}(\mathbf{x}_s^{t} + \mathbf{p}_s).
\end{equation}

\subsection{Vector-Quantized Memory for Storing Learned Prototypes}

During training, PRISM learns a fixed set of compact prototypes representing object-centric features from clean images with complete visual information, and stores them in a vector-quantized memory. During inference, these prototypical slot features are used to correct partial slot representations arising from incomplete observations. We adopt the standard VQ-VAE formulation \cite{oord2017vqvae}, and define a stage-wise codebook as $\mathbf{E}_s$.

Instead of quantizing $\mathbf{k}_{s,\mathrm{grp}}^{t}$ and $\mathbf{v}_{s,\mathrm{grp}}^{t}$ separately, we first concatenate them and map the result into a compact latent space with the encoder $\mathbf{Enc}_s$, which is then quantized using $\mathrm{VQ}$ and decoded back to the grouped representations with the decoder $\mathbf{Dec}_s$:
\begin{equation}
[\hat{\mathbf{k}}_{s,\mathrm{grp}}^{t}, \hat{\mathbf{v}}_{s,\mathrm{grp}}^{t}]
=
Dec_s\!\left(
\mathrm{VQ}\!\left(
Enc_s\!\left([\mathbf{k}_{s,\mathrm{grp}}^{t}, \mathbf{v}_{s,\mathrm{grp}}^{t}]\right);
\mathbf{E}_s
\right)
\right).
\end{equation}

This module serves as a prototype memory over grouped slot representations. During training, it is supervised by two objectives: (1) a reconstruction loss that maps slot representations into the codebook space and reconstructs them via the decoder, and (2) a commitment loss that encourages encoder outputs to remain close to their assigned codebook vectors, stabilizing training and preventing drift in the continuous feature space:
\begin{equation}
\mathcal{L}_{s,\mathrm{vq}}
=
\sum_{t=1}^{T}
\left[
\left\|
[\mathbf{k}_{s,\mathrm{grp}}^{t}, \mathbf{v}_{s,\mathrm{grp}}^{t}]
-
[\hat{\mathbf{k}}_{s,\mathrm{grp}}^{t}, \hat{\mathbf{v}}_{s,\mathrm{grp}}^{t}]
\right\|_2^2
+
\beta
\left\|
\mathbf{z}_s^{t} - \mathrm{sg}[\mathbf{z}_{s,\mathrm{vq}}^{t}]
\right\|_2^2
\right],
\end{equation}
where $\mathrm{sg}[\cdot]$ denotes the stop-gradient operator, $\mathbf{z}_s^{t}$ is the encoded feature vector, and $\mathbf{z}_{s,\mathrm{vq}}^{t}$ is its quantized representation in the codebook (see \textbf{Appendix~Section.~\ref{vq_define}} for formulation).

During inference, grouped representations are passed through the same memory to retrieve nearest prototypes, improving robustness and coherence under incomplete or corrupted observations. The reconstructed grouped key $\hat{\mathbf{k}}_{s,\mathrm{grp}}^{t}$ and value $\hat{\mathbf{v}}_{s,\mathrm{grp}}^{t}$ are then used in the subsequent redistribution step.


\subsection{Redistributing Restored Slot Features back to Their Original Spatial Locations}

Next, $\hat{\mathbf{k}}_{s,\mathrm{grp}}^{t}$ and $\hat{\mathbf{v}}_{s,\mathrm{grp}}^{t}$ are redistributed to their original spatial locations. We use a cross-attention mechanism $\mathrm{CA}$, where position embeddings for each spatial location on the feature map serve as queries, while the restored grouped slot keys and values serve as the key--value pairs:
\begin{equation}
\mathbf{x}_{s,\mathrm{back}}^{t}
=
\mathrm{CA}\!\left(
W_{q,s}\mathbf{p}_s,
\hat{\mathbf{k}}_{s,\mathrm{grp}}^{t}, 
\hat{\mathbf{v}}_{s,\mathrm{grp}}^{t}
\right).
\end{equation}

In this way, $\hat{\mathbf{k}}_{s,\mathrm{grp}}^{t}$ encodes spatially aware information containing “what-to-where”, aligning with positional embeddings to guide the redistribution of $\hat{\mathbf{v}}_{s,\mathrm{grp}}^{t}$ back to the correct spatial locations. Finally, $\hat{\mathbf{k}}_{s,\mathrm{grp}}^{t}$ and $\mathbf{x}_{s,\mathrm{back}}^{t}$ are passed through two independent FFNs with residual connections and MLP updates to produce the slot queries $\mathbf{l}_s^{t+1}$ and input tokens $\mathbf{x}_s^{t+1}$ for the next recurrent step $t+1$.

\subsection{Within-Stage Dynamic Halting for Recurrence Control}

The Adaptive Computation Time (ACT) mechanism dynamically controls when to stop recurrent refinement within each stage by predicting a global halting probability over the patch features. If the accumulated probability exceeds 1, the recurrence stops; otherwise, the process continues until this condition is satisfied. 
ACT allows simpler inputs to terminate early while allocating more computation to harder or more ambiguous cases. This leads to improved computational efficiency and also acts as an implicit regularizer by preventing unnecessary iterative refinement, which helps reduce overfitting.

Formally, each stage is governed by the recurrent refinement function $\Phi_s$, which jointly processes patch features $\mathbf{x}_s^{t}$ and slot queries $\mathbf{l}_s^{t}$, producing updated representations $\mathbf{x}_s^{t+1}$ and $\mathbf{l}_s^{t+1}$ at each step $t$. Since the halting decision depends only on the patch features, we omit slot queries in the notation for simplicity and write the stage output as:
\begin{equation}
\hat{\mathbf{x}}_s = \mathrm{ACT}(\mathbf{x}_s; \Phi_s),
\end{equation}
We use the standard ponder loss for training the ACT module. This loss encourages PRISM to assign higher probability to earlier halting steps, thereby favoring shorter computation when possible. We set the maximum number of recurrent steps to $T_{\max}=5$ for each stage.
\begin{equation}
\mathcal{L}_{\mathrm{ponder},s} = \sum_{t=1}^{T_{\max}} t\,\hat{p}_s^{t}.
\end{equation}



\subsection{Training and Implementation Details}

The overall training objective for PRISM is:
\begin{equation}
\mathcal{L}
=
\mathcal{L}_{\mathrm{cls}}
+
\lambda_{\mathrm{ponder}} \sum_{s=1}^{4}\mathcal{L}_{\mathrm{ponder},s}
+
\lambda_{\mathrm{vq}} \sum_{s=1}^{4}\mathcal{L}_{\mathrm{vq},s},
\end{equation}
where $\lambda_{\mathrm{ponder}} = 0.005$ and $\lambda_{\mathrm{vq}}=0.01$ are the corresponding loss weights, and $\mathcal{L}_{\mathrm{cls}}$ denotes the cross-entropy classification loss.

We train PRISM from scratch for 300 epochs on 4 NVIDIA RTX A6000 GPUs with a total batch size of 1024. Following common ImageNet training practices~\cite{wang2021pvtv2,touvron2021training}, we apply standard data augmentation techniques during training. Optimization is performed using AdamW with an initial learning rate of $1 \times 10^{-3}$, $\beta_1 = 0.9$, and $\beta_2 = 0.999$, together with a cosine learning-rate schedule. Images are resized and cropped to $224 \times 224$ during training, and center cropping is used during validation. 





\section{Experiments}

We evaluate PRISM on image classification, object detection, and semantic segmentation under clean and occluded settings. To benchmark performance, we compare PRISM against strong vision backbones, including PVTv2, BiXT, BiFormer, and DeBiFormer across all tasks. 
All models are trained without occluded images and evaluated on them only at test time to assess out-of-distribution (OOD) robustness. For downstream detection and segmentation tasks, we initialize the backbone with PRISM weights pretrained on ImageNet-1K~\cite{deng2009imagenet} and follow standard fine-tuning protocols \cite{wang2021pvtv2}.

\begin{table*}[t]
\centering
\caption{\textbf{ImageNet-1K classification under clean and occlusion settings.}
We report parameters, FLOPs, and Top-1 accuracy on clean images and five occlusion settings. All experiments have three runs. Standard deviations in Top-1 accuracy for PRISM are reported in brackets (last row). See \textbf{Section.~\ref{exp:classification}} for experimental details. Best results are bolded.
}
\label{tab:classification_occlusion}
\renewcommand{\arraystretch}{1.08}
\resizebox{0.95\textwidth}{!}{
\footnotesize
\begin{tabular}{lcccccccc}
\toprule
\textbf{Model} 
& \textbf{Params} 
& \textbf{FLOPs} 
& \textbf{Clean} 
& \begin{tabular}{c}\textbf{Patch}\\\textbf{Mask}\\\textbf{0.6}\end{tabular}
& \begin{tabular}{c}\textbf{Patch}\\\textbf{Mask}\\\textbf{0.8}\end{tabular}
& \begin{tabular}{c}\textbf{Block}\\\textbf{Mask}\\\textbf{112}\end{tabular}
& \begin{tabular}{c}\textbf{Only}\\\textbf{One}\\\textbf{112}\end{tabular}
& \begin{tabular}{c}\textbf{Only}\\\textbf{One}\\\textbf{56}\end{tabular} \\
\midrule
PVTv2-B1\cite{wang2021pvtv2} & 13M & 2.1G & 78.7 & 53.8 & 39.6 & 74.6 & 62.8 & 27.8 \\
XCiT-T24\cite{ali2021xcit} & 12M & 2.3G & 79.4 & 53.1 & 41.2 & 76.5 & 61.5 & 25.6 \\
BiXT-Ti/16\cite{hiller2024perceiving} & 15M & 1.7G & 80.1 & 55.6 & 40.5 & 77.2 & 64.2 & 28.5 \\
BiFormer-T\cite{zhu2023biformer} & 13M & 2.2G & \textbf{81.4} & 60.4 & 42.1 & 77.2 & 66.4 & 30.4 \\
\midrule
PRISM & 11M & 2.2G & 80.3 & \textbf{69.1} & \textbf{44.6} & \textbf{78.9} & \textbf{69.2} & \textbf{33.8} \\
&     &      & {\scriptsize $\pm$0.12} & {\scriptsize $\pm$0.18} & {\scriptsize $\pm$0.21} & {\scriptsize $\pm$0.10} & {\scriptsize $\pm$0.16} & {\scriptsize $\pm$0.25} \\
\bottomrule
\end{tabular}
}\vspace{-4mm}
\end{table*}

\subsection{Image Classification}\label{exp:classification}
\textbf{Dataset.}
We evaluate image classification on ImageNet-1K~\cite{deng2009imagenet} using the standard training and validation splits.
\textbf{Image Occlusion.}
To assess robustness under incomplete observations, we introduce five occlusion settings on ImageNet-1K images of size $224\times224$: random patch masking with ratios of $0.6$ and $0.8$ (PatchMask${0.6}$ and PatchMask${0.8}$), block occlusion with a region of size $112\times112$ (BlockMask${112}$), and single-visible-patch evaluation retaining only one $112\times112$ or $56\times56$ region (OnlyOne${112}$ and OnlyOne${56}$). These settings simulate varying levels and spatial patterns of missing information.
\textbf{Metrics.}
We report model complexity and performance using number of parameters, FLOPs, and Top-1 accuracy.

\noindent \textbf{Results.}
We report image classification results in \textbf{Table.~\ref{tab:classification_occlusion}}. All models have comparable computational complexity with similar FLOPs. PRISM achieves competitive performance on clean images and substantially stronger robustness under all occlusion settings.

On clean images, PRISM achieves a Top-1 accuracy of $80.3\%$, outperforming PVTv2-B1~\cite{wang2021pvtv2} by $1.6\%$ while using fewer parameters (11M vs.\ 13M). All models experience performance degradation under occlusion, highlighting the difficulty of recognition under incomplete visual evidence. Nevertheless, PRISM consistently surpasses all competitive baselines, with the advantage becoming more pronounced as occlusion severity increases.
For example, under PatchMask$_{0.6}$, PRISM achieves $69.1\%$, outperforming BiFormer-T by $8.7\%$ and PVTv2-B1 by $15.3\%$. Under the more challenging PatchMask$_{0.8}$ setting, PRISM reaches $44.6\%$, compared to $42.1\%$ for BiFormer-T and $39.6\%$ for PVTv2-B1. Similar trends are observed under structured occlusion. For instance, under OnlyOne$_{112}$, PRISM achieves $69.2\%$, exceeding BiFormer-T by $2.8\%$ and PVTv2-B1 by $6.4\%$. These results suggest that PRISM not only improves clean-image recognition, but also maintains more reliable representations when large portions of the image are missing, demonstrating the effectiveness of iterative slot memory for robust visual recognition under incomplete observations.


\subsection{Object Detection and Instance Segmentation}\label{exp:detection}
\textbf{Dataset.}
We evaluate object detection and instance segmentation on the MSCOCO dataset.
\textbf{Implementation Details.}
Following standard transfer learning protocols, PRISM is initialized with weights pretrained on ImageNet-1K and fine-tuned on COCO for detection and instance segmentation. We integrate PRISM into Mask R-CNN and RetinaNet frameworks with a $1\times$ schedule. All models are fine-tuned on 4 NVIDIA RTX A6000 GPUs with a total batch size of 8. We train and evaluate PRISM and all baselines under the same detection framework for fair comparison.
\textbf{Image Occlusion.}
To evaluate robustness under incomplete object observations, we construct occluded COCO validation images by randomly selecting two object instances per image and applying a block mask to each instance. The mask removes either $30\%$ or $50\%$ of the instance area, resulting in two settings (M30 and M50) with different levels of missing visual evidence. This protocol tests whether models can recognize and localize objects when instance-level information is partially occluded.
\textbf{Metrics.}
For RetinaNet, we report bounding-box $\mathrm{mAP}$. For Mask R-CNN, we report $\mathrm{mAP}^{b}$ for bounding-box detection and $\mathrm{mAP}^{m}$ for instance segmentation.


\textbf{Results.}
We report object detection and instance segmentation results under both clean and occluded settings in \textbf{Table.~\ref{tab:downstream_occlusion}(a)}. PRISM achieves competitive clean-image performance while substantially outperforming all baselines under occlusion, despite using significantly fewer parameters.

For RetinaNet, PRISM achieves $43.6$ mAP on clean images, outperforming PVTv2-B1~\cite{wang2021pvtv2} by $2.4$ points and PaCa-Ti~\cite{grainger2023paca} by $1.6$ points, while remaining competitive with DaViT-T~\cite{ding2022davit} using roughly half the number of parameters. Under occlusion, the advantage of PRISM becomes more pronounced. PRISM achieves $38.8$ mAP under M30 and $34.2$ mAP under M50, surpassing DaViT-T by $1.9$ and $2.7$ points, respectively, indicating stronger robustness when object regions are partially missing.

For Mask R-CNN, PRISM also achieves strong detection and instance segmentation performance despite having the fewest parameters among all models. On clean images, PRISM achieves $44.5$ $\mathrm{mAP}^{b}$ and $41.1$ $\mathrm{mAP}^{m}$, matching DaViT-T in mask accuracy while using approximately $40\%$ fewer parameters. Under occlusion, the gains become larger. For example, under M30, PRISM reaches $40.6$ $\mathrm{mAP}^{b}$ and $36.9$ $\mathrm{mAP}^{m}$, improving over DaViT-T by $2.0$ and $1.6$ points, respectively. Under the more challenging M50 setting, PRISM further achieves $37.2$ $\mathrm{mAP}^{b}$ and $33.7$ $\mathrm{mAP}^{m}$, outperforming DaViT-T by $4.2$ and $3.9$ points. These results suggest that PRISM improves not only classification robustness, but also object localization and instance-level segmentation under incomplete observations.

\subsection{Semantic Segmentation}\label{exp:segmentation}
\textbf{Dataset.}
We evaluate semantic segmentation on ADE20K.
\textbf{Implementation Details.}
Following standard transfer learning protocols, PRISM is initialized with weights pretrained on ImageNet-1K and fine-tuned on ADE20K. We select a standard semantic segmentation framework and fine-tune all models with the same framework for fair comparison.
\textbf{Image Occlusion.}
To evaluate robustness under incomplete observations, we construct occluded ADE20K validation images by randomly masking $30\%$ of image patches. This setting removes scattered spatial evidence and tests whether models can maintain dense semantic predictions when parts of the scene are missing.
\textbf{Metrics.}
We report mean Intersection-over-Union (mIoU), which measures the average overlap between predicted and ground-truth semantic regions across all object classes.

\textbf{Results.}
Results for PRISM and baseline models under clean and occluded settings are reported in \textbf{Table.~\ref{tab:downstream_occlusion}(b)}. PRISM achieves the best semantic segmentation performance under both clean and occluded settings.
On clean images, PRISM achieves an mIoU of $44.8$, outperforming PVTv2-B1~\cite{wang2021pvtv2} by $2.3$ points and BiXT-Ti/16~\cite{hiller2024perceiving} by $5.6$ points, despite using slightly fewer parameters. More importantly, PRISM maintains the best performance under occlusion with $39.7$ mIoU, while PVTv2-B1 and BiXT-Ti/16 drop to $34.8$ and $30.5$, respectively. This corresponds to a gain of $4.9$ points over PVTv2-B1 under random patch masking.
The strong performance on dense prediction tasks suggests that PRISM preserves spatially coherent object-centric representations, enabling more reliable semantic grounding even when parts of the scene are missing.


\begin{table*}[t]
\centering
\caption{\textbf{Downstream task performance under clean and occlusion settings.}
(a) MSCOCO object detection and instance segmentation. Parameter count is reported in Column 2. In Row 2, C denotes the clean setting, and M30/M50 denote occlusion settings. Superscripts $b$ and $m$ indicate the metric type: for Mask R-CNN, bounding box detection ($\mathrm{mAP}^{b}$) and mask segmentation ($\mathrm{mAP}^{m}$) are reported, while no superscript denotes $\mathrm{mAP}$ with RetinaNet. See \textbf{Section.~\ref{exp:detection}} for experimental details.
(b) ADE20K semantic segmentation. Parameter count (Params.) and FLOPs are reported in Columns 2 and 3. C and Occ.\ denote the clean and occlusion settings. Model performance is reported in mIoU. See \textbf{Section.~\ref{exp:segmentation}} for experimental details. Best are bolded. (`--`) indicates that the model checkpoint is not officially provided and their result is therefore unavailable.}
\label{tab:downstream_occlusion}
\small
\setlength{\tabcolsep}{2.8pt}
\renewcommand{\arraystretch}{1.02}

\resizebox{\textwidth}{!}{
\begin{tabular}{@{}c@{\hspace{0.045\textwidth}}c@{}}
\textbf{(a) MSCOCO detection and instance segmentation}
&
\textbf{(b) ADE20K semantic segmentation}
\\[1.5mm]

\begin{tabular}{lccccccccccc}
\toprule
\multirow{2}{*}{\textbf{Model}}
& \multicolumn{4}{c}{\textbf{RetinaNet}}
& \multicolumn{7}{c}{\textbf{Mask R-CNN}} \\
\cmidrule(lr){2-5} \cmidrule(lr){6-12}
& \textbf{Param}
& \textbf{C}
& \textbf{M30}
& \textbf{M50}
& \textbf{Param}
& \textbf{C$^b$}
& \textbf{C$^m$}
& \textbf{M30$^b$}
& \textbf{M30$^m$}
& \textbf{M50$^b$}
& \textbf{M50$^m$} \\
\midrule
PVTv2-B1\cite{wang2021pvtv2} & 23.8M & 41.2 & 32.8 & 27.6 & 33.7M & 41.8 & 38.8 & 34.1 & 31.1 & 28.7 & 25.9 \\
PaCa-Ti\cite{grainger2023paca} & -- & -- & -- & -- & 32.0M & 43.3 & 39.6 & 37.5 & 34.7 & 31.8 & 28.7 \\
DaViT-T\cite{ding2022davit} & 38.5M & \textbf{44.0} & 36.9 & 31.5 & 47.8M & \textbf{45.0} & \textbf{41.1} & 38.6 & 35.3 & 33.0 & 29.8 \\
\midrule
PRISM  & 20.5M & 43.6 & \textbf{38.8} & \textbf{34.2} & 31.4M & 44.5 & 40.8 & \textbf{40.6} & \textbf{36.9} & \textbf{37.2} & \textbf{33.7} \\
\bottomrule
\end{tabular}
&
\begin{tabular}{lcccc}
\toprule
\multirow{2}{*}{\textbf{Model}}
& \multirow{2}{*}{\textbf{Param.}}
& \multirow{2}{*}{\textbf{FLOPs}}
& \multicolumn{2}{c}{\textbf{mIoU}} \\
\cmidrule(lr){4-5}
&
&
& \textbf{C}
& \textbf{Occ.} \\
\midrule
XCiT-T12\cite{ali2021xcit} & 8M & - & 39.9 & 32.2 \\
PVTv2-B1\cite{wang2021pvtv2}   & 17.8M & 34.2G & 42.5 & 34.8 \\
BiXT-Ti/16\cite{hiller2024perceiving} & 19M & 31.8G & 39.2 & 30.5 \\
\midrule
PRISM      & 15.5M & 35.6G & \textbf{44.8} & \textbf{39.7} \\
\bottomrule
\end{tabular}
\end{tabular}
}
\vspace{-4mm}
\end{table*}

\subsection{Ablation Studies on PRISM}

\begin{table}[t]
\centering
\caption{\textbf{Ablation studies of PRISM on ImageNet-1K classification.} We report Top-1 accuracy on clean images and three occlusion conditions. Our default PRISM uses a maximum of 5 halting steps in (a), includes full VQ memory in (b), and employs dynamic halting (Dyn) in (c). Best are bolded.}\vspace{2mm}
\footnotesize
\begin{tabular}{lccccccccccc}
\hline
 & \multicolumn{3}{c}{\shortstack{(a) Max halting steps\\during training}} 
 & \multicolumn{2}{c}{\shortstack{(b) Remove VQ\\during inference}} 
 & \multicolumn{6}{c}{\shortstack{(c) Fixed recurrent steps\\during inference}} \\
\cline{2-4} \cline{5-6} \cline{7-12}
Model Variants 
& $4$ & $5$ & $6$ 
& w/o VQ & Full
& 1 & 2 & 3 & 4 & 5 & Dyn \\
\hline
Clean 
& 80.0 & 80.3 & \textbf{80.4} 
& 80.2 & \textbf{80.3} 
& 78.0 & 79.2 & 80.2 & 80.3 & 80.3 & \textbf{80.3} \\

PatchMask0.8 
& 43.7 & 44.6 & \textbf{44.8} 
& 37.8 & \textbf{44.6} 
& 31.4 & 38.9 & 43.1 & 44.0 & 44.2 & \textbf{44.6} \\

BlockMask112 
& 78.4 & 78.9 & \textbf{79.0} 
& 75.6 & \textbf{78.9} 
& 73.5 & 76.8 & 78.5 & 78.8 & 78.8 & \textbf{78.9} \\

OnlyOne112 
& 68.1 & 69.2 & \textbf{69.4} 
& 62.7 & \textbf{69.2} 
& 58.3 & 64.1 & 68.0 & 68.8 & 69.0 & \textbf{69.2} \\
\hline
\end{tabular}\vspace{-4mm}
\label{tab:ablation}
\end{table}

We conduct ablation studies of PRISM on ImageNet-1K classification under clean images and a subset of occlusion conditions described in \textbf{Section.~\ref{exp:classification}}. Results in Top-1 accuracy are reported in \textbf{Table.~\ref{tab:ablation}}.
\noindent \textbf{Effect of maximum halting steps.}
Varying the maximum number of halting steps has limited impact on clean accuracy but significantly affects robustness under occlusion. Reducing the maximum steps from $5$ to $4$ slightly decreases clean accuracy ($80.3 \rightarrow 80.0$), but leads to larger drops under occlusion (e.g., $69.2 \rightarrow 68.1$ on OnlyOne112). Increasing the limit from $5$ to $6$ yields only marginal gains (e.g., $44.6 \rightarrow 44.8$ on PatchMask0.8 and $69.2 \rightarrow 69.4$ on OnlyOne112), suggesting that $5$ steps are sufficient for most inputs.
\noindent \textbf{Effect of VQ memory.}
Removing VQ memory at inference has negligible effect on clean accuracy ($80.3 \rightarrow 80.2$), but causes substantial degradation under occlusion, e.g., $44.6 \rightarrow 37.8$ on PatchMask0.8 and $69.2 \rightarrow 62.7$ on OnlyOne112. This highlights the importance of prototype-based restoration when visual evidence is incomplete.
\noindent \textbf{Effect of dynamic halting.}
Increasing the number of fixed recurrent steps improves performance (e.g., $31.4 \rightarrow 43.1$ on PatchMask0.8 and $58.3 \rightarrow 68.0$ on OnlyOne112 from $T{=}1$ to $T{=}3$), but saturates after $4$--$5$ steps. Dynamic halting achieves $44.6$ on PatchMask0.8 and $69.2$ on OnlyOne112, outperforming fixed-step inference ($T{=}5$), demonstrating its ability to adapt computation to input difficulty for improved robustness and efficiency.

\subsection{Model Analysis}

\begin{figure*}[t]
    \centering
    \includegraphics[width=0.9\linewidth]{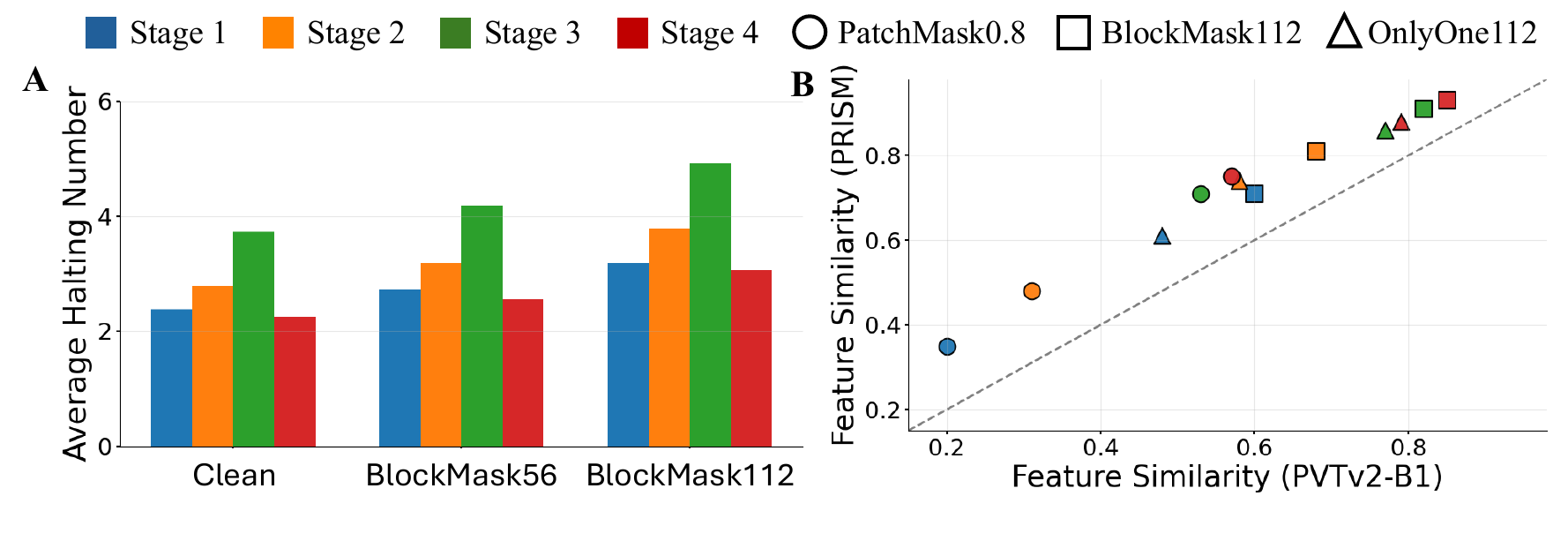}\vspace{-2mm}
    \caption{\textbf{Analysis of the halting dynamics of PRISM in image classification.}
\textbf{A.} Average halting steps at each stage under clean and two occlusion settings. Colors denote different stages. 
\textbf{B.} Feature similarity between clean and occluded inputs, comparing PVTv2-B1 (x-axis) and PRISM (y-axis) across stages and occlusion settings. The dashed line denotes the diagonal. Each point corresponds to one stage (color) and one occlusion setting (marker). See \textbf{Section.~\ref{exp:classification}} for details on occlusion settings.
    }\vspace{-4mm}
    \label{fig:analysis}
\end{figure*}

To probe the halting dynamics of PRISM, we analyze inference-time halting frequency across four stages of the visual hierarchy (\textbf{Figure.~\ref{fig:analysis}A}), as well as feature similarity between clean and occluded inputs (\textbf{Figure.~\ref{fig:analysis}B}). Also see an analysis of training-time halting frequencies in \textbf{Appendix~Section.~\ref{halting_training}}.
Inference-time halting frequency under occlusion in \textbf{Figure.~\ref{fig:analysis}A} shows that PRISM allocates more refinement steps as occlusion severity increases. For example, in Stage 3, the average halting steps increase from 3.73 (clean) to 4.18 (BlockMask56) and 4.92 (BlockMask112), indicating that more computation is required under harder conditions. Across the four stages of the visual hierarchy, Stage 3 consistently requires the most refinement, while Stage 4 remains lower, suggesting that intermediate stages are most responsible for feature restoration.
Representation consistency results in \textbf{Figure.~\ref{fig:analysis}B} show that PRISM preserves higher feature similarity than PVTv2-B1 across all stages and occlusion settings. For instance, under PatchMask0.8, Stage 3 similarity increases from 0.53 (PVTv2-B1) to 0.71 (PRISM). This suggests that adaptive recurrence and the prototypical representations in VQ memory help keep occluded features closer to their clean counterparts, improving model robustness and classification accuracy.

\section{Discussion}
We present PRISM, a pyramid vision architecture that reasons over images through iterative slot memory. Instead of a single feed-forward pass, PRISM iteratively organizes patch features into slot-level entities, restores them via a vector-quantized prototype memory, and adapts the number of recurrent steps to input complexity via dynamic halting. Across image classification, object detection, and semantic segmentation, PRISM achieves competitive performance on standard benchmarks while showing stronger robustness under occlusion, suggesting that compositional abstraction, memory-based restoration, and adaptive recurrence are effective for robust visual recognition. Despite these benefits, the inference cost of PRISM may increase for challenging inputs requiring more refinement steps. Moreover, our evaluation primarily focuses on synthetic occlusions, and further validation under real-world occlusion scenarios remains important.


\bibliography{example_paper}
\bibliographystyle{abbrv}

\newpage
\appendix
\renewcommand{\thesection}{A\arabic{section}}
\renewcommand{\thefigure}{A\arabic{figure}}
\renewcommand{\thetable}{A\arabic{table}}
\setcounter{figure}{0}
\setcounter{section}{0}
\setcounter{table}{0}

\section{Preliminaries}\label{supp:preliminaries}

We introduce the basic operators used in our model, including attention-based feature aggregation, vector quantization, and adaptive computation.

\subsection{Cross-Attention (CA) and Slot-Attention (SA)}
Given query, key, and value matrices $\mathbf{Q} \in \mathbb{R}^{N_q \times D}$, $\mathbf{K} \in \mathbb{R}^{N_k \times D}$, and $\mathbf{V} \in \mathbb{R}^{N_k \times D}$, the dot product $\mathbf{A}$ between $\mathbf{Q}$ and $\mathbf{K}$ is:
\begin{equation}
\mathbf{A}=
\frac{\mathbf{Q}\mathbf{K}^{\top}}{\sqrt{D}}
\in
\mathbb{R}^{N_q \times N_k}
\end{equation}
The output of a cross-attention operation, denoted by $\mathrm{CA}(\mathbf{Q}, \mathbf{K}, \mathbf{V})$, is defined as:
\begin{equation}
\texttt{attn}_{i,j}=
\frac{\exp^{A_{i,j}}}{\sum_{N_k}\exp^{A_{i,j}}}, \mathrm{CA}(\mathbf{Q}, \mathbf{K}, \mathbf{V})=\texttt{attn} \times V
\end{equation}

For Slot Attention\cite{locatello2020slotattention}, we similarly write the grouping operation as $\mathrm{SA}(\mathbf{Q}, \mathbf{K}, \mathbf{V})$. Its output is defined as:
\begin{equation}
\texttt{attn}_{i,j}=
\frac{\exp^{A_{i,j}}}{\sum_{N_q}\exp^{A_{i,j}}}, \mathbf{W_{i,j}}=\frac{\texttt{attn}_{i,j}}{\sum_{N_k}\texttt{attn}_{i,j}}, \mathrm{SA}(\mathbf{Q}, \mathbf{K}, \mathbf{V})=\mathbf{W} \times V \label{equ:equAttSA}
\end{equation}
Here, $\mathrm{attn}_{i,j}$ normalizes over the slot dimension so that slots compete for each token, while $W_{i,j}$ further normalizes over the token dimension before aggregation.

\subsection{Vector Quantization (VQ)}\label{vq_define}
Given an input feature matrix $\mathbf{Z} \in \mathbb{R}^{N \times D}$ and a codebook $\mathbf{E}=\{\mathbf{e}_m\}_{m=1}^{M}$ with $\mathbf{e}_m \in \mathbb{R}^{D}$, vector quantization maps each feature $\mathbf{z}_i$ to its nearest code:
\begin{equation}
m_i=\arg\min_{m}\|\mathbf{z}_i-\mathbf{e}_m\|_2^2,
\qquad
\mathbf{z}_{q,i}=\mathbf{e}_{m_i}.
\end{equation}
The quantized output is denoted by $\mathbf{Z}_q=\mathrm{VQ}(\mathbf{Z};\mathbf{E})$. We adopt the EMA-style codebook update in VQ-VAE ~\cite{oord2017vqvae}, where each code is updated by the moving average of the features assigned to it. In this way, the codebook tracks the centers of the assigned features during training.

\subsection{Adaptive Computation Time (ACT)}
Given token features $\mathbf{X}$, we define the dynamic halting process as $\hat{\mathbf{X}}=\mathrm{ACT}(\mathbf{X}; \Phi)$, where $\mathbf{X}^{t} \in \mathbb{R}^{N \times D}$ denotes the token features at iteration $t$, and $\hat{\mathbf{X}} \in \mathbb{R}^{N \times D}$ denotes the final aggregated output. The features are updated recurrently by an updating function $\Phi$:
\begin{equation}
\mathbf{X}^{t+1}=\Phi(\mathbf{X}^{t}).
\end{equation}
Instead of predicting halting probabilities for individual tokens, an improved version of predicting single global halting probability for the whole token set at each iteration is introduced. Specifically, a global feature is first computed:
\begin{equation}
\bar{\mathbf{x}}^{t}=\frac{1}{N}\sum_{n=1}^{N}\mathbf{x}_n^{t},
\end{equation}
and then predict the halting variables by
\begin{equation}
p^{t}=\sigma(\mathrm{FC}(\bar{\mathbf{x}}^{t})), \,
T=\min\left\{t:\sum_{t'=1}^{t}p^{t'} \ge 1-\epsilon\right\}, \,
R=1-\sum_{t=1}^{T-1}p^{t},
\end{equation}
where $p^{t} \in (0,1)$ is the global halting probability at iteration $t$, $T$ is the halting step, and $R$ is the remainder at the halting step. We set the hyperparmaeter $\epsilon=0.01$ to improve numerical stability and enable smooth halting behavior by introducing a slightly relaxed stopping threshold, which helps stabilize the learning of the halting mechanism. The final output is computed as a weighted aggregation over recurrent states, where intermediate steps use their halting probabilities as weights, the final halting step uses the remainder, and steps after $T$ receive zero weight. The aggregation over time steps is defined as
\begin{equation}
\hat{p}^{t}=
\left\{
\begin{array}{ll}
p^{t}, & t<T,\\
R, & t=T,\\
0, & t>T,
\end{array}
\right.
\qquad
\hat{\mathbf{X}}=\sum_{t=1}^{T_{\max}}\hat{p}^{t}\mathbf{X}^{t},
\end{equation}
where $T_{\max}$ is the maximum number of iterations.

\section{Stage-wise Halting Dynamics during Training}\label{halting_training}

\begin{figure}[t]
    \centering
    \includegraphics[width=0.8\linewidth]{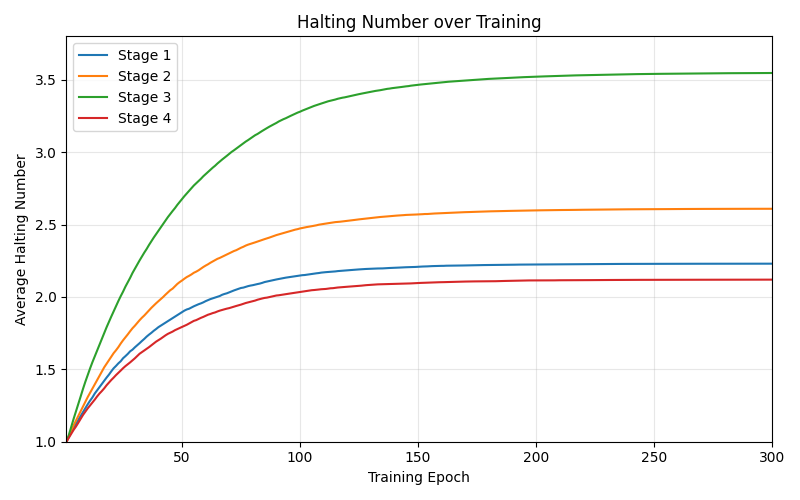}
    \caption{\textbf{Halting dynamics of PRISM.} Average halting numbers as a function of training epochs, showing how the model adaptively allocates recurrent refinement steps during training.}
    \label{fig:halting_training}
\end{figure}

As shown in \textbf{Figure.~\ref{fig:halting_training}}, the halting numbers of all four stages increase rapidly in the early phase of training and become nearly stable after roughly the first one-third of optimization. This suggests that the adaptive computation policy is learned relatively early, after which the model mainly focuses on improving the quality of its representations rather than substantially changing how much computation is allocated. In other words, the network first discovers \emph{how much} iterative refinement each stage should perform, and then uses the remaining training process to improve \emph{what} is computed within this recurrent budget. 

A second important observation is that the learned halting numbers are strongly stage-dependent in the visual hierarchy of PRISM. Stage 3 converges to the largest halting number, followed by Stage 2, while Stage 1 and Stage 4 remain lower. This pattern is meaningful rather than incidental. The early stage mainly operates on relatively local patterns, where recurrent refinement is useful but still limited by the low-level nature of the features. The deepest stage (Stage 4), in contrast, already works on highly abstract representations, so fewer additional refinement steps are needed once the semantics have largely stabilized. The middle stages (Stages 2 and 3) therefore become the main locus of recurrent computation, where local evidence and global structure must be integrated into coherent object-level representations. This stage-wise allocation indicates that the halting mechanism does not collapse to a trivial constant-depth solution, but instead learns a structured computation pattern aligned with the semantic roles of the pyramid stages.

\end{document}